# Exploiting T-norms for Deep Learning in Autonomous Driving

Mihaela C. Stoian[1,*], Eleonora Giunchiglia[2] and Thomas Lukasiewicz[2,1]

[1]*Department of Computer Science, University of Oxford, UK*
[2]*Institute of Logic and Computation, TU Wien, Austria*

### Abstract
Deep learning has been at the core of the autonomous driving field development, due to the neural networks' success in finding patterns in raw data and turning them into accurate predictions. Moreover, recent neuro-symbolic works have shown that incorporating the available background knowledge about the problem at hand in the loss function via t-norms can further improve the deep learning models' performance. However, t-norm-based losses may have very high memory requirements and, thus, they may be impossible to apply in complex application domains like autonomous driving. In this paper, we show how it is possible to define memory-efficient t-norm-based losses, allowing for exploiting t-norms for the task of event detection in autonomous driving. We conduct an extensive experimental analysis on the ROAD-R dataset and show (i) that our proposal can be implemented and run on GPUs with less than 25 GiB of available memory, while standard t-norm-based losses are estimated to require more than 100 GiB, far exceeding the amount of memory normally available, (ii) that t-norm-based losses improve performance, especially when limited labelled data are available, and (iii) that t-norm-based losses can further improve performance when exploited on both labelled and unlabelled data.

## 1. Introduction

Deep learning has been at the core of the autonomous driving field development [1, 2], due to the neural networks' success in finding patterns in raw data and turning them into accurate predictions. However, existing self-driving vehicle systems are very limited in their capabilities [3], with many of the obstacles in reaching a fully autonomous system being rooted in the underlying neural models' own caveats, such as the inherent data greediness and the impossibility of incorporating background knowledge about the problem at hand. Recently, neuro-symbolic methods emerged as a way to integrate background knowledge within the neural networks' topology (see, e.g., [4, 5, 6]) and/or loss function (see, e.g., [7, 8, 9]), with a large number of them highlighting a positive impact particularly in scenarios where little annotated data is available (see, e.g., [10, 11, 12]). A popular method to include background knowledge expressed as logical constraints into neural networks consists in relaxing the constraints using t-norms

---

and incorporating them in the loss function [7, 8, 13]. Such a t-norm-based loss function is not only intuitive, but it also has been shown to improve neural networks' performance on a range of different tasks (including event detection in autonomous driving [14]), especially when limited data are available. However, t-norm-based losses may have very high memory requirements, and thus they may be impossible to apply when considering complex application domains like event detection in autonomous driving.

In this paper, we show how it is possible to define memory-efficient t-norm-based losses, allowing for exploiting t-norms for the task of event detection in autonomous driving. We conduct an extensive experimental analysis on the ROAD-R dataset [14] and show that our proposal can be implemented and run on GPUs with less than 25 GiB of available memory, while standard t-norm-based losses are estimated to require more than 100 GiB, far exceeding the amount of memory normally available. Then, we train our state-of-the-art event detection model using different percentages of labelled training data (i.e., 10%, 20%, 50%, 75%, and 100%), and we show that while t-norm-based losses can improve the performance of the models in all cases, they are particularly helpful when data are scarce. Indeed, our models yield an improvement of up to 1.85% and 3.95% when using, 10% and 20% of the labelled training data, respectively. Finally, we investigate the behaviour of the t-norm-based loss when only 10% annotated data are available, along with 10% unlabelled data, and find that applying the t-norm-based loss on both labelled and unlabelled data after a warm-up training phase leads to further improvements, i.e., up to 2.75% w.r.t. the fully supervised baseline.

## 2. Preliminaries

An *event detection problem* $\mathcal{P}$ is a pair $(\mathcal{A}, \mathcal{X})$, where $\mathcal{A}$ is a finite set of labels, and $\mathcal{X}$ is set of pairs $(\mathbf{X}, \mathcal{Y})$ where:

1. $\mathbf{X} \in \mathbb{R}^{3 \times W \times H}$ is the tensor associated with each frame in the video. $W$ (resp., $H$) represents the width (resp., height) of each frame, while 3 is the number of channels used in the RGB encoding,
2. $\mathcal{Y}$ is the ground truth of $\mathbf{X}$ comprising a set of pairs $(\mathbf{b}, \mathcal{L})$, where $\mathbf{b} \in \mathbb{R}^4$ represents the coordinates of a *bounding box*, i.e., a rectangle marking the position of an agent in the frame, while $\mathcal{L}$ represents the set of labels associated with $\mathbf{b}$.

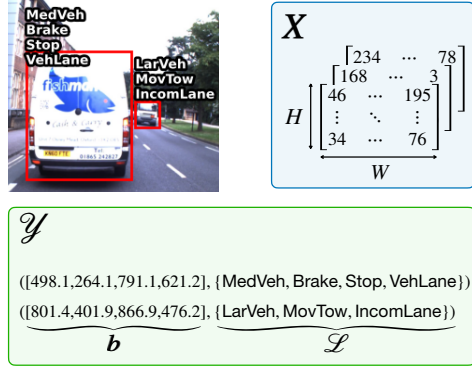

**Figure 1:** Visual example of a data point.

Figure 1 illustrates an example of a data point $\mathbf{X}$ and its ground truth $\mathcal{Y}$. A *model m* for $\mathcal{P}$ takes as input a sequence of video frames and, for each input frame $\mathbf{X}$, its *outputs* are the pairs $(\hat{\mathbf{b}}, \hat{\mathbf{y}})$, where each $\hat{\mathbf{b}} \in \mathbb{R}^4$ represents a predicted bounding box, and $\hat{\mathbf{y}} \in [0, 1]^{|\mathcal{A}|}$ represents the confidence of the model regarding which labels can be associated with $\hat{\mathbf{b}}$. Given an output $(\hat{\mathbf{b}}, \hat{\mathbf{y}})$, a *prediction* is then defined as the pair $(\hat{\mathbf{b}}, \hat{\mathcal{L}})$, where $\hat{\mathcal{L}}$ is the set of labels associated with $\hat{\mathbf{b}}$; a label $A$ is associated with $\hat{\mathbf{b}}$ if $\hat{\mathbf{y}}_A \geq \theta$, where $\theta$ is a user-defined threshold. To predict the bounding boxes in each frame, standard off-the-shelf event detection models use *anchor boxes*, which are predefined boxes of varying sizes at different locations within the frame. For each frame, the model defines $D$ anchor boxes whose positions are fixed, and then the position of each predicted bounding box is computed as the offset from one anchor box.

An *event detection problem with propositional logic constraints* $(\mathscr{P}, \Pi)$ consists of an event detection problem $\mathscr{P}$ and a finite set of constraints $\Pi$, expressed over the set $\mathscr{A}$ of labels in $\mathscr{P}$. We assume w.l.o.g. that the constraints are given as a set of *clauses*, each of the form:

$$l_1 \vee l_2 \vee \cdots \vee l_n, \qquad (1)$$

where every $l_i$ is a *literal*, i.e., is either a label $A \in \mathscr{A}$ or its negation, $\neg A$, for $i \in \{1, \ldots, n\}$. Intuitively, (1) expresses the fact that the model should always predict at least one of the literals in the clause, i.e., in $\{l_1, \ldots, l_n\}$. We assume that in any clause, a label occurs either positively or negatively at most once. We say that a label $A$ *occurs positively* (resp., *negatively*) in the clause (1) if there is a literal $l$ in (1) such that $l = A$ (resp., $l = \neg A$), and that $A$ *occurs* in (1) if $A$ occurs either positively or negatively in (1).

## 3. Memory-efficient t-norm-based loss

Inspired by [13, 7], we added a new regularization term to the localisation and classification losses to express the degree of the logical constraints satisfaction. Since our constraints are all of the form (1), we can easily convert each of them into a form containing only negations and conjunctions (i.e., $l_1 \vee l_2 \vee \cdots \vee l_n \equiv \neg(\neg l_1 \wedge \neg l_2 \wedge \cdots \wedge \neg l_n)$). We can then relax:

1. the conjunction using different t-norms [15]. A *t-norm* is a function $T : [0,1]^2 \to [0,1]$ such that for every $a, b, c \in [0,1]$:

$$T(a,b) = T(b,a), \quad T(a,1) = a, \quad T(a,0) = 0, \quad T(a, T(b,c)) = T(T(a,b), c),$$
$$a \leq b \to T(a,c) \leq T(b,c).$$

2. the negation using *strong negation*, which, given $a \in [0,1]$, is defined as $1 - a$.

The above operation is equivalent to directly relaxing the disjunction using the appropriate t-conorm. In the first two columns of Table 1, we summarise the most used t-norms together with their respective t-conorms. We now show how to implement t-norm-based loss functions first in the standard way, and then in a memory-efficient way using sparse tensors.

Let $(\mathscr{P}, \Pi)$ be an event detection problem with propositional logic constraints. We can then express $\Pi$ using two matrices $\boldsymbol{C}^+$ and $\boldsymbol{C}^-$, both of size $|\Pi| \times |\mathscr{A}|$, such that $C^+_{ij} = 1$ (resp., $C^-_{ij} = 1$), if the $j$-th label appears positively (resp., negatively) in the $i$th constraint, and $C^+_{ij} = 0$ (resp., $C^-_{ij} = 0$), otherwise. We call $\boldsymbol{C}^+$ (resp., $\boldsymbol{C}^-$) the *positive* (resp., *negative*) *constraints matrix*. Let $m$ be an event detection model for $\Pi$ using $D$ anchor boxes for each frame. For each input frame, $m$ outputs the *prediction matrix* $\boldsymbol{P}$ of size $D \times |\mathscr{A}|$. We call $\boldsymbol{P}$ the *prediction matrix*. Given $\boldsymbol{P}$ and $\boldsymbol{C}$, our goal is to compute the degree of satisfaction of each constraint for each output, which can be compactly expressed as a matrix $\boldsymbol{G}$ of size $D \times |\Pi|$. Ultimately, we want to use $\boldsymbol{G}$ to compute the frame-wise logic-based regularisation term in the loss, and we do this by defining:

$$L_{logic}(\boldsymbol{G}) = 1 - \frac{1}{D}\frac{1}{|\Pi|} \sum_{ij} G_{ij}. \qquad (2)$$

**Standard Approach.** Let $\hat{P}$ be the tensor obtained by stacking $|\Pi|$ times the matrix $P$ along its first dimension. Let $\hat{C}^+$ and $\hat{C}^-$ be the tensors obtained by stacking $D$ times the matrices $C^+$ and $C^-$ along their second dimension. We then obtain three tensors all of size $D \times |\Pi| \times |\mathscr{A}|$. We can then choose the desired t-conorm and compute the matrix goal $G$ as:

$$G = \text{t-conorm}([(\hat{P} \odot \hat{C}^+) + (\hat{C}^- - \hat{P} \odot \hat{C}^-)], \dim = 3),$$

where $\odot$ is the Hadamard product, and given a generic tensor $Q$ of size $p \times q \times s$, t-conorm($Q$, dim = 3) returns a matrix of size $p \times q$ whose element at position $(i, j)$ is equal to the value of the t-conorm computed over the third dimension, e.g., if we choose the Gödel t-conorm, we obtain t-conorm($Q$, dim = 3) = $\max(Q_{ij1}, Q_{ij2}, \ldots, Q_{ijs})$.

**Example 3.1.** *Let $(\mathscr{P}, \Pi)$ be an event detection problem with propositional logic constraints such that $\mathscr{A} = \{Car, Moving, Stopped\}$ and $\Pi = \{\neg Moving \vee Car, \neg Moving \vee \neg Stopped\}$. Then, our positive and negative constraints matrices, assuming labels are numbered as listed in $\mathscr{A}$, would be:*

$$C^+ = \begin{bmatrix} 1 & 0 & 0 \\ 0 & 0 & 0 \end{bmatrix} \qquad C^- = \begin{bmatrix} 0 & 1 & 0 \\ 0 & 1 & 1 \end{bmatrix}.$$

*Let $m$ be a model for $(\mathscr{P}, \Pi)$ using 3 anchor boxes. Given a prediction matrix $P$ as below, and supposing we use the Gödel t-conorm, $G$ is equal to:*

$$P = \begin{bmatrix} 0.1 & 0.7 & 0.3 \\ 0.9 & 0.9 & 0.2 \\ 0.4 & 0.9 & 0.9 \end{bmatrix} \qquad G = \begin{bmatrix} 0.3 & 0.7 \\ 0.9 & 0.8 \\ 0.4 & 0.1 \end{bmatrix}.$$

The problem with this approach is that it requires working with dense 3-dimensional matrices, inducing a large computational overhead and making the method unfeasible, especially for application domains like autonomous driving. For example, given $|\Pi| = 200$ constraints, $|\mathscr{A}| = 50$ labels, and a model generating $D = 55K$ anchor boxes per frame (a common number for event detection problems) and taking as input sequences of 10 frames at a time, storing a single matrix of size $D \times |\Pi| \times |\mathscr{A}|$ requires about 20 GiB ($550000 \times 200 \times 50 \times 4$ bytes). Moreover, the standard approach works with 5 matrices of this size to compute the t-norm loss in the forward pass and then backpropagate through it. Excluding any other memory allocation needed for storing the input and intermediate outputs, just computing the loss and backpropagating through it would take 100 GiB, exceeding the memory space limits of even the largest GPU available today (i.e., the NVIDIA A100 Tensor Core GPU having 80 GiB RAM [16]). Notice that this computation is done for a single frame, however, normally deep learning models for event detection are trained using batches of 4/8 elements, each comprising 4 to 32 frames in sequence. It is thus impossible to use the above standard dense representation to train event detection models.

**Sparse Matrix Representation Approach.** Our solution mostly relies on the intuition that in practice most of the constraints are written over a subset of the available labels $\mathscr{A}$, and that this subset is usually much smaller than $\mathscr{A}$. For example, in our experimental analysis, we will see that although there are 41 labels available in ROAD-R, the longest constraint is written over

**Table 1**

From left to right: (i) t-norm definitions, (ii) respective t-conorm definitions, and (iii-iv) operations to update $G$ on the grounds of the chosen t-conorm. Given two matrices, max (resp., min) represent the element-wise operation taking the maximum (resp., minimum) between two elements at the same position in the two input matrices. To simplify the notation, in the last two columns, we used 1 to refer to the matrix of ones of appropriate size.

|  | T-norm | T-conorm | Operation to update $G_{j_A^+}$ | Operation to update $G_{j_A^-}$ |
|---|---|---|---|---|
| Gödel | $\min(a,b)$ | $\max(a,b)$ | $\max(G_{j_A^+}, P_A \cdot \mathbb{1}_{|j_A^+|}^\top)$ | $\max(G_{j_A^-}, 1 - P_A \cdot \mathbb{1}_{|j_A^-|}^\top)$ |
| Łukasiewicz | $\max(a+b-1, 0)$ | $\min(a+b, 1)$ | $\min(G_{j_A^+} + P_A \cdot \mathbb{1}_{|j_A^+|}^\top, 1)$ | $\min(G_{j_A^-} + 1 - P_A \cdot \mathbb{1}_{|j_A^-|}^\top, 1)$ |
| Product | $a \cdot b$ | $1 - (1-a)(1-b)$ | $1 - (1 - G_{j_A^+}) \odot (1 - P_A \cdot \mathbb{1}_{|j_A^+|}^\top)$ | $1 - (1 - G_{j_A^-}) \odot (P_A \cdot \mathbb{1}_{|j_A^-|}^\top)$ |

just 15 labels. As a result, $C^+$ and $C^-$ contain mostly zeros. Hence, we designed a method to capture the logic-based loss that makes use of this sparsity property and ultimately avoids the high computational costs induced by the 3D matrices, operating only on 2D matrices.

Given $\Pi$, we associate with each constraint an index, and then we define the set of sequences $\mathcal{J}^+ = \{j_A^+ : A \in \mathcal{A}\}$, where $j_A^+$ is the sequence of indices of the constraints in which $A$ occurs positively. Analogously, we define $\mathcal{J}^- = \{j_A^- : A \in \mathcal{A}\}$, where $j_A^-$ is the sequence of indexes of the constraints in which $A$ occurs negatively. Once we know which constraints each label occurs in, we can instantiate the goal matrix $G$ to the identity element of the disjunction, iterate through the labels in $\mathcal{A}$, and for each label $A \in \mathcal{A}$ update, according to the values in $P$, all the columns of $G$ associated with constraints where $A$ occurs. More specifically, we set $G = \mathbf{0}_{D \times |\Pi|}$, where $\mathbf{0}_{D \times |\Pi|}$ is a matrix of zeros of size $D \times |\Pi|$, and then, for each label $A \in \mathcal{A}$, we determine:

$$G_{j_A^+} \longleftarrow \text{t-conorm}(G_{j_A^+}, P_A \cdot \mathbb{1}_{|j_A^+|}^\top) \qquad G_{j_A^-} \longleftarrow \text{t-conorm}(G_{j_A^-}, 1 - P_A \cdot \mathbb{1}_{|j_A^-|}^\top),$$

where (i) $G_{j_A^+}$ (resp., $G_{j_A^-}$) selects the columns of $G$ associated with constraints where $A$ occurs positively (resp., negatively), (ii) $P_A$ corresponds to the column of $P$ associated with the label $A$, (iii) $\mathbb{1}_n$ indicates the unit column vector with $n$ elements, and (iv) t-conorm returns the pairwise t-conorm, i.e., given two matrices $W, Z$ of the same size, t-conorm$(W, Z)_{ij}$ = t-conorm$(W_{ij}, Z_{ij})$. Finally, given $G$, we compute $L_{logic}(G)$ as defined in Equation 2.

**Example 3.2 (Example 3.1, cont'd).** *Let $(\mathcal{P}, \Pi)$ be the problem in Example 3.1 and assume that we associate with the constraint $(\neg Moving \lor Car)$ index 0, and with $(\neg Moving \lor \neg Stopped)$ index 1. Let $\epsilon$ denote the empty sequence. Then, the sequences associated with each label are:*

$$j_{Car}^+ = (0), \quad j_{Moving}^+ = \epsilon, \quad j_{Stopped}^+ = \epsilon, \quad j_{Car}^- = \epsilon, \quad j_{Moving}^- = (0, 1), \quad j_{Stopped}^- = (1).$$

*Suppose that we use the Gödel t-conorm, then after having initialized $G = \mathbf{0}_{3 \times 2}$, we start updating it from the label Car:*

$$G_{j_{Car}^+} = \max\left(\begin{bmatrix} 0 \\ 0 \\ 0 \end{bmatrix}, \begin{bmatrix} 0.1 \\ 0.9 \\ 0.4 \end{bmatrix}\right) = \begin{bmatrix} 0.1 \\ 0.9 \\ 0.4 \end{bmatrix} \quad \text{and thus} \quad G = \begin{bmatrix} 0.1 & 0 \\ 0.9 & 0 \\ 0.4 & 0 \end{bmatrix}$$

*Since $j_{Car}^- = \epsilon$, we do not need to further update $G$ for Car. We then consider the label Moving and update it according to $j_{Moving}^-$ (as $j_{Moving}^+ = \epsilon$):*

$$G_{j_{Moving}^-} = \max\left(\begin{bmatrix} 0.1 & 0 \\ 0.9 & 0 \\ 0.4 & 0 \end{bmatrix}, 1 - \begin{bmatrix} 0.7 & 0.7 \\ 0.9 & 0.9 \\ 0.9 & 0.9 \end{bmatrix}\right) = \begin{bmatrix} 0.3 & 0.3 \\ 0.9 & 0.1 \\ 0.4 & 0.1 \end{bmatrix} \quad \text{and thus} \quad G = \begin{bmatrix} 0.3 & 0.3 \\ 0.9 & 0.1 \\ 0.4 & 0.1 \end{bmatrix}.$$

*Finally, we consider the label Stopped, and perform the last update:*

$$G_{j_{Stopped}^-} = \max\left(\begin{bmatrix}0.3\\0.1\\0.1\end{bmatrix}, 1 - \begin{bmatrix}0.3\\0.2\\0.9\end{bmatrix}\right) = \begin{bmatrix}0.7\\0.8\\0.1\end{bmatrix} \quad \text{and thus} \quad G = \begin{bmatrix}0.3 & 0.7\\0.9 & 0.8\\0.4 & 0.1\end{bmatrix}.$$

## 4. Experimental Analysis

We tested our t-norm loss on the task of event detection for autonomous driving, where the goal is to assign to each detected bounding box in each video a subset of labels—including one agent label, and a subset of the action and location labels. To this end, we used the recently introduced dataset for autonomous driving, ROAD-R [14], which extends the ROAD dataset [17] with 243 manually annotated constraints, provided in disjunctive normal form, as shown in Table 5 from Appendix A. The dataset contains 22 videos, each ~8 minutes long, annotated with tubelets/tubes that link a sequence of bounding boxes in time. Each bounding box is annotated with a subset of the 41 labels available (listed in Table 4 from Appendix A). We used the available training partition for training the models and, for reproducibility purposes, we reported our results on the validation dataset, as the test set is not publicly available. For our experiments, we used the 3D-RetinaNet [17] detector with a ResNet50 [18] backbone combined with a Random Connectivity Gated Recurrent Unit (RCGRU) [19] for temporal feature learning, which we chose based on its high performance in [14]. We set a weight of 10 for the t-norm loss term and use sequences of 8 frames as input.

**Memory assessment.** To assess the efficiency of our method for computing the t-norm loss w.r.t. how much GPU memory is allocated during training the models, we compared it to the standard implementation of the t-norm-based loss. We used a Titan RTX GPU with 24 GiB of RAM for training models for 50 iterations on ROAD-R, while using different numbers of constraints. Note that, while most of the constraints in ROAD-R contain only two labels, to allow for a fair comparison with the standard implementation, we selected constraints with a different numbers of labels. For reference, in each iteration, the number of anchors $D$ was of about 67K. Figure 2 shows that our method significantly reduced the memory costs, making it possible to use t-norm-based losses on our dataset, where the number of constraints and data points per batch are both large. Using the standard implementation supports at most 40 constraints; this is 203 constraints less than the ones in ROAD-R.

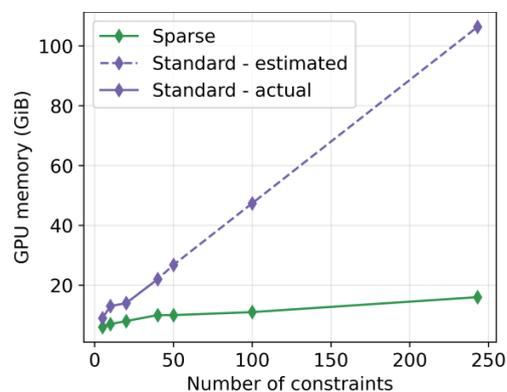

**Figure 2:** Comparison between the standard approach and ours in terms of GPU memory allocated when using different number of constraints. Each point on the continuous (resp., dashed) line corresponds to an actual observation (resp., estimate).

**Table 2**

Comparison between our models with different t-norm-based losses and the baseline models when varying the percentage of labelled data, as indicated on top of each column.

|  | 10% | 20% | 50% | 75% | 100% |
|---|---|---|---|---|---|
| Baseline | 24.49 | 26.81 | 31.56 | 33.57 | 33.39 |
| Gödel | **26.34** (+1.85) | **30.76** (+3.95) | 32.71 (+1.15) | **34.39** (+0.82) | 32.16 (−1.23) |
| Łukasiewicz | 26.24 (+1.75) | 29.13 (+2.32) | **34.07** (+2.51) | 33.89 (+0.32) | **34.42** (+1.03) |
| Product | 24.53 (+0.04) | 26.96 (+0.15) | 31.66 (+0.10) | 34.06 (+0.49) | 33.34 (−0.05) |

**Results.** We first investigated how our memory-efficient t-norm-based loss performs in a fully-supervised scenario. To this end, we tested our method with three different t-norm losses (i.e., Gödel, Łukasiewicz, and Product) using 10%, 20%, 50%, 75% and 100% of the available annotated ROAD-R data, and training for 110, 70, 45, 30, and 30 epochs, respectively. We used a learning rate of 0.0041 for all models, but for 100% labelled data, we dropped it at epochs 18 and 25 by a factor of 10, as in [14]. We always computed the t-norm-based losses w.r.t. all of the 243 constraints from ROAD-R. For evaluation, we used the frame-wise mean average precision (f-mAP) metric, computed by taking the mean average precision at a fixed intersection-over-union threshold of 0.5 over each frame for each class and then averaging these per-class scores, and reported the result at the best epoch.

Table 2 summarises the results, from which we first observe that t-norm-based losses always improve the baseline performance, except when using 100% labelled data, where only the Łukasiewicz t-norm outperforms the baseline—this being in line with the result on 100% labelled data from our previous work[1] [14]. We also notice that unlike the Gödel and Łukasiewicz t-norms, in most cases, the Product t-norm brings negligible improvements, if any. Lastly, as expected, integrating background knowledge (via t-norm loss) in the neural models helps more when little data are available. Indeed, our models yield an improvement of up to 1.85% and 3.95% when using, 10% and 20% of the labelled training data, respectively.

The last observation led us to investigate whether in our setting also holds the known result that background knowledge helps when unlabelled data are available (see, e.g., [20, 12]). To this end, we trained models where we applied the t-norm based losses on 10% labelled and 10% unlabelled data. As shown in the second column of Table 3, neither the Gödel nor the Łukasiewicz t-norm were particularly helpful w.r.t. their fully-supervised performance. Surprisingly, the Product t-norm improved its performance instead, now surpassing the baseline. Since the added unlabelled data were not really helpful in two out of three cases, we analysed the losses at the beginning of the training and hypothesised that it would be beneficial to introduce a warm-up training phase, during which the t-norm loss would be inactive, and after which the unlabelled data would be added and the t-norm loss activated. As expected, the results from the last column of Table 3 consistently improve previous performances of all t-norm based losses, with the Product t-norm giving the highest result (of 27.24 f-mAP) w.r.t. the baseline (of 24.49 f-mAP).

**Table 3:** The **best**- and <u>worst</u>-performing models across fully-supervised models and models using unlabelled data, with or without warm-up. All models here used 10% labelled data for training. The models in the last two columns used also 10% unlabelled data during the training phase.

|  | Fully-sup. | With unlabelled data | |
|---|---|---|---|
|  | - | - | Warm-up |
| Gödel | <u>26.34</u> | 26.38 | **26.76** |
| Łukasiewicz | 26.24 | <u>25.48</u> | **26.75** |
| Product | <u>24.53</u> | 25.79 | **27.24** |

---

[1]These results are in line with those obtained in our work [14], where an early and less optimised version (nevertheless, still capable of handling all 243 constraints) of this implementation of the t-norm-based loss had been deployed but not described.

## 5. Related Work

Neuro-symbolic works have proposed ways to embed available background knowledge by either embedding it into the topology and/or into the loss. In the former category, we find works that build a constrained layer on top of a neural network, such as Coherent-by-Construction Network (CCN) [5], MultiplexNet [21], and Semantic Probabilistic Layer (SPL) [6], all of these approaches being able to guarantee the satisfaction of hard constraints under certain conditions. Having a similar goal, NESTER [22] proposes another end-to-end approach by imposing soft and hard constraints via a program applied on the outputs. Yet another recent work is Iterative Local Refinement (ILR) [23], which proposed an analytic way of integrating t-norm-based functions as neural network layers to refine the predictions in a differentiable manner.

The other main line of work comprises methods that relax the constraints and integrate them into the neural networks' loss, directly relating to our method. Early work on semantic based regularisation (SBR) [24] on kernel machines led to the development of ways to map the constraints into the neural networks' loss according to the t-norm operations [7, 12, 25, 26]. However, among other issues highlighted in [27, 28], one problem occurring in these approaches is that they are syntax-dependent. To address this, Semantic Loss [9] and DL2 [11] introduced syntax-independent loss functions. Another work, by Ahmed et al. [29], integrates logic into the standard entropy regularisation [30] term of the loss. While the most recent work, by Li et al. [31], explores another problem with the previous approaches, namely, that models tend to settle on the easier solutions that satisfy the constraints, and proposes a way to enforce the model to fully explore the available knowledge. While many such works proved to be particularly helpful when little annotated data are available [10, 11, 12, 32, 33, 20], they have been designed for small and/or synthetic datasets and would not scale to complex scenarios. For a complete survey on how to incorporate logical constraints in deep learning, see [34].

## 6. Conclusion

In this paper, we formalise an approach for computing a memory-efficient t-norm-based loss to equip neural networks with background knowledge logical constraints. We show that, unlike standard implementations of t-norm-based losses, our method can be applied in resource-intensive scenarios, such as event detection for autonomous driving. On the ROAD-R dataset, we test our t-norm-based loss on different amounts of labelled data showing that the t-norms indeed help in boosting the performance of state-of-the-art models, and we also present an effective way to use the t-norms in presence of unlabelled data.

For future work, we plan on conducting a study into how the loss can help in a semi-supervised scenario, with varying amounts of unlabelled data added during the training and using the warm-up phase that we found helpful here for reducing the runtime strain brought by integrating background knowledge into neural networks. Another research direction could be a study on the effect of using different intervals of warm-up, before introducing the t-norm loss and applying it on the unlabelled data.

## Acknowledgments

Mihaela C. Stoian is supported by the EPSRC under the grant EP/T517811/1. This work was also supported by the Alan Turing Institute under the EPSRC grant EP/N510129/1, by the AXA Research Fund, by the EPSRC grant EP/R013667/1, and by the EU TAILOR grant. We also acknowledge the use of the EPSRC-funded Tier 2 facility JADE (EP/P020275/1) and GPU computing support by Scan Computers International Ltd.

# A. Appendix: the ROAD-R dataset

| ID | Agent | ID | Action | ID | Location |
|----|-------|----|--------|----|----------|
| 0 | Pedestrian | 10 | Move away | 29 | AV lane |
| 1 | Car | 11 | Move towards | 30 | Outgoing lane |
| 2 | Cyclist | 12 | Move | 31 | Outgoing cycle lane |
| 3 | Motorbike | 13 | Brake | 32 | Incoming lane |
| 4 | Medium vehicle | 14 | Stop | 33 | Incoming cycle lane |
| 5 | Large vehicle | 15 | Indicating left | 34 | Pavement |
| 6 | Bus | 16 | Indicating right | 35 | Left pavement |
| 7 | Emergency vehicle | 17 | Hazard lights on | 36 | Right pavement |
| 8 | AV traffic light | 18 | Turn left | 37 | Junction |
| 9 | Other traffic light | 19 | Turn right | 38 | Crossing location |
|  |  | 20 | Overtake | 39 | Bus stop |
|  |  | 21 | Wait to cross | 40 | Parking |
|  |  | 22 | Cross road from left |  |  |
|  |  | 23 | Cross road from right |  |  |
|  |  | 24 | Crossing |  |  |
|  |  | 25 | Push object |  |  |
|  |  | 26 | Red traffic light |  |  |
|  |  | 27 | Amber traffic light |  |  |
|  |  | 28 | Green traffic light |  |  |

**Table 4**
The labels (with their ID's) from the ROAD-R dataset used in our experiments.

| Logical constraints | Descriptions in natural language |
|---------------------|----------------------------------|
| {not Mobike, not Bus} | A motorbike cannot be a bus |
| {not TL, not TurLft} | A traffic light cannot turn left |
| {not Wait2X, not Ovtak} | An agent cannot wait to cross and overtake |
| {Ped, not PushObj} | If an agent pushes an object then it is a pedestrian |
| {PushObj, not Ped, MovAway, MovTow, Mov, Stop, TurLft, TurRht, Wait2X, XingFmLft, XingFmRht, Xing} | A pedestrian can only push objects, move away, etc. |
| {VehLane, OutgoLane, OutgoCycLane, IncomLane, IncomCycLane, Pav, LftPav, RhtPav, Jun, XingLoc, BusStop, Parking, TL, OthTL} | Every agent but traffic lights must have a position |

**Table 5**
Examples of logical constraints and their natural language explanations from Tables 9-11 of [14].